\RequirePackage{fix-cm}

\documentclass[twocolumn]{svjour3}          
\smartqed  
\usepackage{graphicx}

\usepackage{booktabs,multirow}
\usepackage{array}
\usepackage{amsmath}
\usepackage{enumitem}
\usepackage{bm} 
\usepackage{authblk} 

\begin{document}
\title{TSNet:A Two-stage  Network for Image Dehazing with Multi-scale Fusion and Adaptive Learning
}


\author{Xiaolin Gong \textsuperscript{1,2}         \and
       Zehan Zheng \textsuperscript{1}   \and Heyuan Du \textsuperscript{1} 
}


\institute{
	Xiaolin Gong \at
	gxl@tju.edu.cn          
	\and
	Zehan Zheng \at
	zzh07@tju.edu.cn          
	\and
	Heyuan Du \at
	2022232168@tju.edu.cn  
	\and
	1 School of Microelectronics, Tianjin University, Tianjin 300072, China 
	\at
	2 Tianjin Key Laboratory of Imaging and Sensing Microelectronic Technology, Tianjin University, Tianjin, 300072, China 
}

\date{Received: date / Accepted: date}

\maketitle

\begin{abstract}
	Image dehazing has been a popular topic of research for a long time. Previous deep learning-based image dehazing methods have failed to achieve satisfactory dehazing effects on both synthetic datasets and real-world datasets, exhibiting poor generalization. Moreover, single-stage networks often result in many regions with artifacts and color distortion in output images. To address these issues, this paper proposes a two-stage image dehazing network called TSNet, mainly consisting of the multi-scale fusion module (MSFM) and the adaptive learning module (ALM). Specifically, MSFM and ALM enhance the generalization of TSNet. The MSFM can obtain large receptive fields at multiple scales and integrate features at different frequencies to reduce the differences between inputs and learning objectives. The ALM can actively learn of regions of interest in images and restore texture details more effectively. Additionally, TSNet is designed as a two-stage network, where the first-stage network performs image dehazing, and the second-stage network is employed to improve issues such as artifacts and color distortion present in the results of the first-stage network. We also change the learning objective from ground truth images to opposite fog maps, which improves the learning efficiency of TSNet. Extensive experiments demonstrate that TSNet exhibits superior dehazing performance on both synthetic and real-world datasets compared to previous state-of-the-art methods. The related code is released at https://github.com/zzhlovexuexi/TSNet
\end{abstract}

\section{Introduction}
\label{intro}
In adverse weather conditions, the atmosphere contains various particulates, leading to a significant degradation in the image quality captured by imaging devices \cite{1}. For instance, images exhibit severe color distortion and loss of details in hazy weather, posing significant challenges for both human and machine operations \cite{2}. Therefore, how to restore dehazed images from blurry ones has been one of the popular research topics in recent years.

Earlier, an atmospheric scattering model was widely applied in image dehazing methods based on prior knowledge \cite{3}.
\begin{equation}
	\mathrm{\textit{I}}(\mathrm{\textit{x}})=\mathrm{\textit{J}}(\mathrm{\textit{x}}) \mathrm{\textit{t}}(\mathrm{\textit{x}})+\mathrm{\textit{A}}(1-\mathrm{\textit{t}}(\mathrm{\textit{x}}))\label{eq:1}
\end{equation}
Where \textit{I} is the original hazy image, \textit{J} is the latent haze-free image, \textit{A} is the global atmospheric light and \textit{t} is the medium transmission map, indicating the degree of attenuation as light passes through the atmospheric particulates, and its value is between 0 and 1.

\begin{figure}
    \hspace{-0.38cm}
	\includegraphics[height=6.5cm]{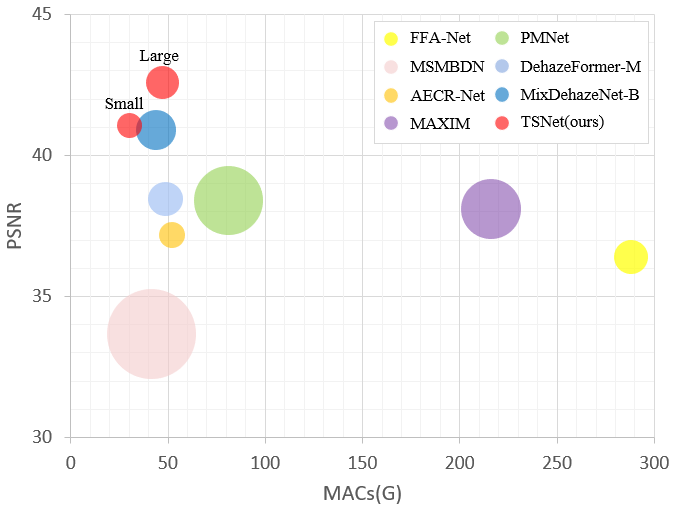}
	\caption{The results of TSNet compared with other state-of-the-art methods on the RESIDE-IN dataset. The size of the circle represents \#Param}
	\label{fig:1}
\end{figure}

If both \textit{t}(\textit{x}) and \textit{A} are known, the latent dehazed image can be obtained using Eq.(\ref{eq:1}). Therefore, researchers have devoted efforts to obtaining \textit{t}(\textit{x}) and \textit{A} using prior knowledge \cite{4,5,6,7}. However, the estimated values of \textit{t}(\textit{x}) and \textit{A} deviate significantly from the actual values when real-world scenes are complex \cite{29}, leading to a sharp decline in the performance of prior knowledge-based dehazing methods.

In recent years, the rapid development of deep learning has also brought new opportunities to the field of image dehazing, with an increasing number of researchers utilizing learning-based methods for image dehazing research \cite{12,13,14,15,16,17,18,19,20,21,22,23,24,25}. Learning-based dehazing methods can handle more complex real-world scenarios with the help of large-scale datasets. Despite significant progress in previous research, limitations still exist, (1) In single-stage networks, output images often have significant color distortion and blocky artifacts, and the optimization of outputs is not possible. (2) Previous networks display significant variations in dehazing performance across datasets and cannot achieve satisfactory dehazing results simultaneously on synthetic and real-world datasets.

To address the aforementioned issues, we designed a two-stage image dehazing network named TSNet inspired by \cite{8,9,10,11}. The structure of TSNet is based on the U-net \cite{30}. It mainly consists of multi-scale fusion modules (MSFM) and adaptive learning modules (ALM). MSFM and ALM enhance the generalization of TSNet, MSFM uses a multi-scale parallel large convolution kernel module (MSPLCK) to obtain large receptive fields \cite{25}, establishing a more complex spatial structure. It then uses an implicit frequency feature enhancement module (IFFE) to integrate different frequency features, reducing the gap in frequency information between inputs and learning objectives. Benefiting from the deformable convolutional network DCN \cite{31}, ALM can dynamically adjust the sampling range of convolution kernels, making the sampling points more closely adhere to the sampled objects \cite{47}. Therefore, it achieves a better recovery of details for objects. Additionally, TSNet is divided into two stages, the first stage of the network is used for image dehazing. The second-stage network aimed at optimizing the results of the first-stage network to reduce artifacts and perform color correction. The two stages of TSNet are trained independently, and their weights are stored separately. Internally, the second-stage network can be viewed as a simplified version of the first-stage network, eliminating the need to design new modules specifically for the second stage. Unlike previous networks where the learning objective is ground truth images, we change the learning objective to the opposite fog maps. This change is made to enhance the learning efficiency of the TSNet. Fig. \ref{fig:1} shows TSNet compared with previous methods on the RESIDE-IN dataset. Our small model is lower in overhead but better in performance than the previous state-of-the-art method, MixDehazeNet-B \cite{25}. The contributions of this paper are summarized as follows:

\begin{itemize}[label=$\bullet$]
	\item We proposed a two-stage dehazing network called TSNet. The first-stage network performs image dehazing, and the second-stage network is employed to optimize the results of the first-stage network, effectively reducing the artifacts and color distortion commonly present in the outputs of single-stage networks.
\end{itemize}

\begin{itemize}[label=$\bullet$]
	\item We designed a multi-scale fusion module. The multi-scale fusion module can obtain multi-scale large receptive fields and integrate features at different frequencies to reduce the differences between inputs and learning objectives.
\end{itemize}

\begin{itemize}[label=$\bullet$]
	\item An adaptive learning module is created that improves the ability to recover texture details by dynamically adjusting the sampling range of convolution kernels. Additionally, it can be employed as a universal module by other networks to enhance dehazing performance.
\end{itemize}

\begin{itemize}[label=$\bullet$]
	\item By changing the learning learning objective of TSNet from ground truth images to the opposite fog maps, the learning efficiency of the network is enhanced.
\end{itemize}

\begin{figure*}
	\centering
	\includegraphics[width=1\textwidth]{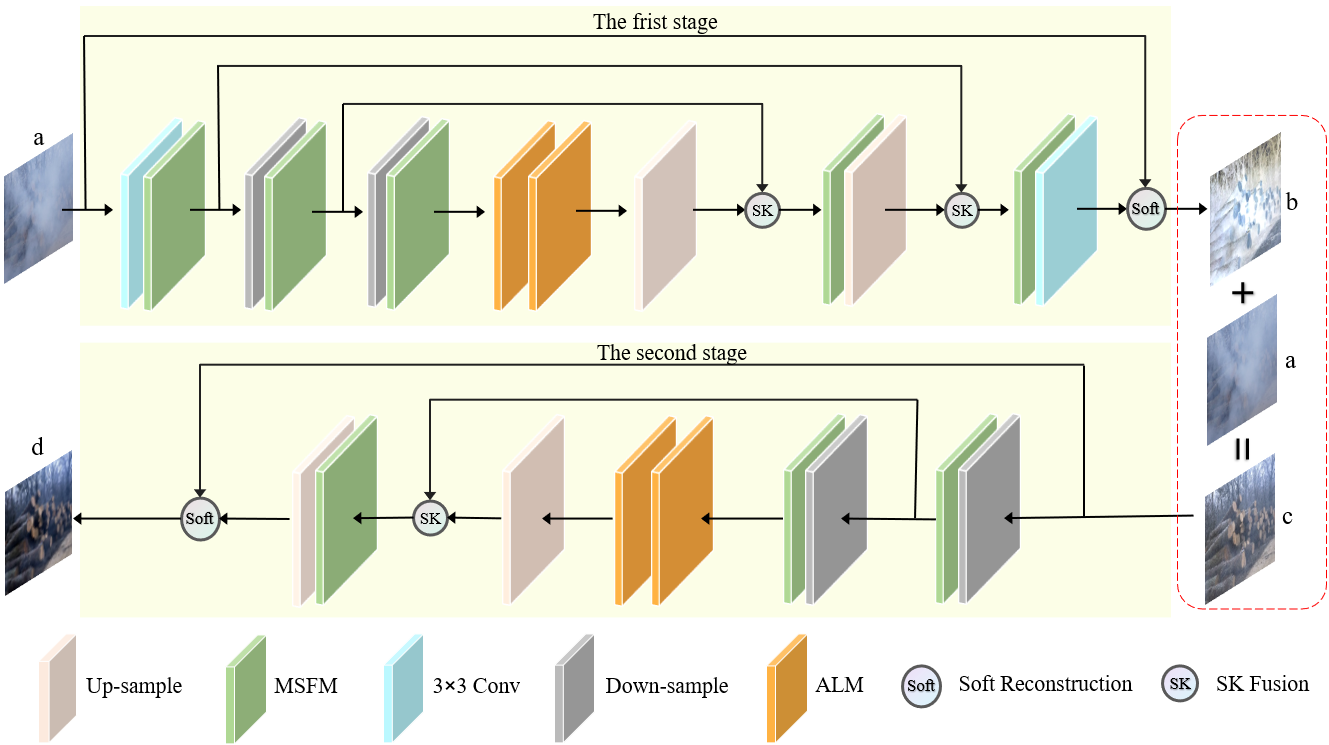}
	\caption{The internal structure of TSNet. The dashed box represents the process of changing the learning objective, `a' is the hazy image, `b' is the opposite fog map, `c' is the dehazed image obtained from the first-stage network, and `d' is the final dehazed image of TSNet}
	\label{fig:2}
\end{figure*}

\section{Related Works}
There are two categories of impressive methods in image dehazing, methods based on prior knowledge \cite{4,5,6,7} and methods based on deep learning \cite{12,13,14,15,16,17,18,19,20,21,22,23,24,25}.

\textbf{Prior-based image dehazing}: Early researchers obtained valuable statistical information and prior knowledge by observing a large number of haze images and ground truth images, which they then applied to image dehazing. DCP \cite{4} proposed that the minimum value of the image channel in the local haze area always approaches 0. It then estimated \textit{A} and \textit{t}(\textit{x}) and used a soft matting interpolation algorithm to obtain dehazed images. CAP \cite{6} observed that when a ground truth image was affected by haze, the saturation decreased and brightness increased, which manifested as color attenuation. So it proposed a linear formula to estimate \textit{t}(\textit{x}). NLD \cite{7} discovered that the RGB values of a ground truth image form a cluster in the RGB space. However, in a hazy image, the RGB values form a line called haze-lines in the RGB space. This characteristic was utilized to obtain \textit{t}(\textit{x}). Prior-based image dehazing methods have low time complexity and minimal hardware requirements \cite{46}, so they have generated a lot of related research. But in reality, the value of the scattering coefficient in \textit{t}(\textit{x}) varies at different spatial locations \cite{32}. However, dehazing methods based on prior knowledge usually set it as a constant, assuming that the particulates in the air are uniformly distributed. So these methods may not effectively remove haze in complex scenes.
\begin{figure*}
	\centering
	\includegraphics[width=0.8\textwidth]{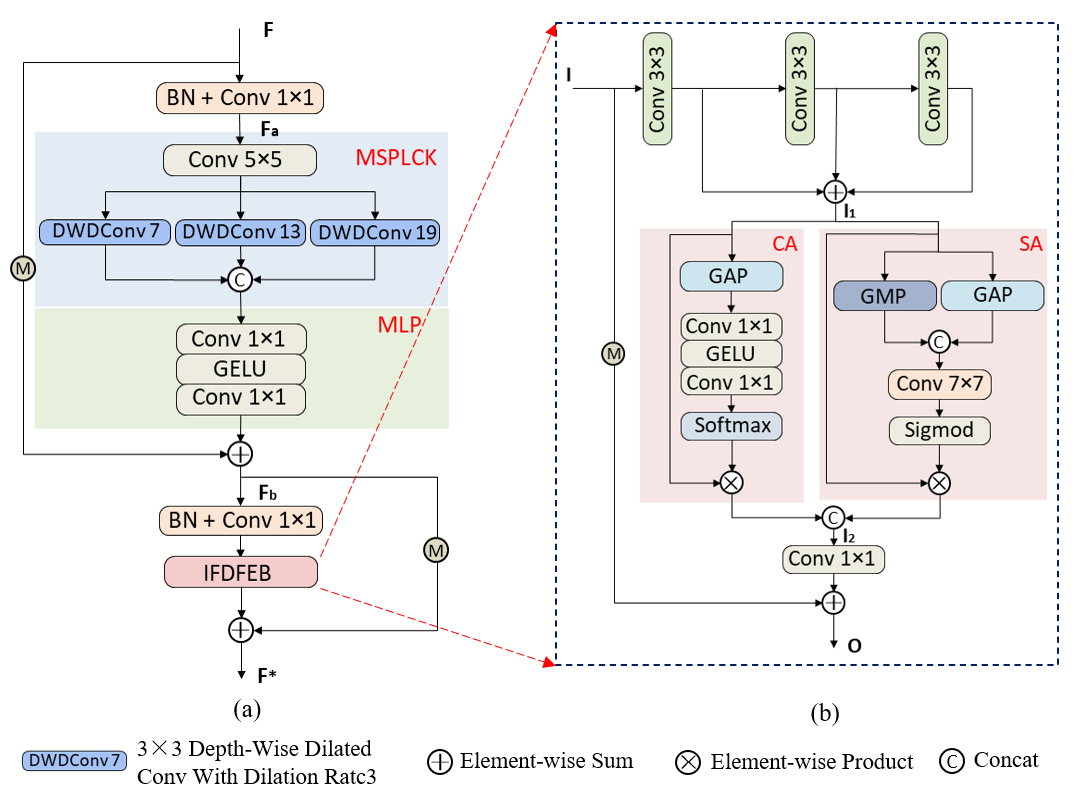}
	\caption{(a) is the structure of the multi-scale fusion module, mainly consisting of a multi-scale parallel convolutional kernel module and an implicit frequency feature enhancement module. (b) is the internal structure of the implicit frequency feature enhancement module}
	\label{fig:3}
\end{figure*}

\textbf{Learning-based image dehazing}: The development of deep learning has brought new opportunities to image dehazing. The early network DehazeNet \cite{14} used neural networks to learn \textit{t}(\textit{x}) and \textit{A}, which were then fed into the atmospheric scattering model to obtain dehazed images. In the same year, MSCNN \cite{12} designed a coarse-scale network to learn \textit{t}(\textit{x}), which was then optimized with a fine-scale network. AOD \cite{13} redefined the atmospheric scattering model to avoid estimating \textit{t}(\textit{x}) and \textit{A} separately, reducing error accumulation. GridDehazeNet \cite{28} employed an end-to-end network to directly output dehazed images, surpassing previous parameter-learning methods. MSBDN \cite{18} designed a dense feature fusion module using the back-projection feedback scheme to address the problem of missing spatial information in high-resolution features in the U-Net architecture. FFA \cite{19} further improved dehazing effectiveness by utilizing skip connections and attention mechanisms. AECR-Net \cite{24} introduced a contrastive loss into the loss function, serving as a general approach to enhance dehazing performance. MAXIM \cite{20} proposed a multi-axis MLP architecture, where the stacked encoder-decoder framework made the network more flexible. PMNet \cite{23} proposed a novel separable hybrid attention module to encode haze density and a density map to model the uneven distribution of the haze, addressing the problem of modeling real-world haze degradation. DehazeFormer \cite{21} utilized the Swin Transformer and improved several key designs, achieving good results in the field of image dehazing. MixDehazeNet \cite{25} designed a parallel attention module and a multi-scale parallel large convolutional kernel module, surpassing all previous methods on the RESIDE-IN dataset \cite{33}. Although deep learning-based methods have achieved satisfactory dehazing results, these methods often suffer from poor generalization and lack consideration for further optimization of the dehazed images, resulting in noticeable artifacts and color distortion in the outputs.

\section{TSNet}
This section primarily introduces the structure of TSNet, as well as the learning objectives and loss functions of the first-stage and the second-stage network. As shown in Fig. \ref{fig:2}, TSNet is divided into two stages, the first stage is a modified 5-stage U-Net and the second stage is a modified 3-stage U-Net \cite{30}, whose convolutional blocks are replaced by our multi-scale fusion modules. Additionally, the adaptive learning module is deployed before the first up-sampling in each stage of the TSNet. SK Fusion and Soft Reconstruction \cite{21} are employed to fuse shallow and deep features for better learning effects.

\subsection{Multi-scale Fusion Module}
The internal structure of the multi-scale fusion module is shown in Fig. \ref{fig:3}. It contains a multi-scale parallel large convolution kernel module (MSPLCK) and an implicit frequency feature enhancement module (IFFE). The MSPLCK acquires multi-scale receptive fields, which provide a better understanding of the global context and structure of the scene \cite{25}. The IFFE obtains features at different frequencies and then integrates these features to reduce the differences in frequency information between inputs and learning objectives.

First, we use BatchNorm to normalize the input \textbf{F}, preventing gradient vanishing and gradient exploding during training, and reducing TSNet's sensitivity to initial values \cite{37}.
\begin{equation}
    \textbf{F}_\textbf{a}={\mathrm{Conv}}_{{1} \times {1}}\left({{\mathrm{Batchnorm}}}(\textbf{F})\right)
\end{equation}
Where {Conv}\textsubscript{{\textit{i}}×{\textit{i}}} represents a convolution operation with a kernel size of {\textit{i}}×\textit{\textit{i}}.

Then the MSPLCK is employed with dilated convolution kernels of different sizes to obtain multi-scale receptive fields \cite{25}. It allows the network to better handle global structures. Concat operation is used to concatenate multiple tensors along a specified dimension. Subsequently, an MLP filters and retains more useful features. The outputs of MLP are then added to M({\textbf{F}}) to reduce information loss during propagation.
\begin{equation}
	  \mathrm{\textbf{F}}_\mathrm{\textbf{b}}=\mathrm{MLP}(\mathrm{MSPLCK}(\mathrm{\textbf{F}}_\mathrm{\textbf{a}}))+\mathrm{M}(\mathrm{\textbf{F}}) 
\end{equation}
Where $\mathrm{M}(\mathrm{\textbf{F}})$ is \textit{k}·{\textbf{F}}, representing assigning a weight \textit{k} to feature \textbf{F}. The value of \textit{k} can be dynamically adjusted to an appropriate value during the training process ($0 \leq \textit{k}\leq 1$).

Then the IFFE is applied to reduce the differences in frequency information between the inputs and the learning objectives. Finally, the outputs are added to M(\textbf{F}\textsubscript{\textbf{b}}) to obtain the output \textbf{F*}, as shown in Eq. \ref{eq:4}.
\begin{equation}
	\begin{split}
		\mathrm{\textbf{F*}} = & \mathrm{IFFE}\left(({\mathrm{Conv}}_{{1} \times {1}}( \right. \\
		& \left. \mathrm{Batchnorm}(\mathrm{\textbf{F}}_\mathrm{\textbf{b}})) +\mathrm{M}(\mathrm{\textbf{F}}_\mathrm{\textbf{b}}) \right)\label{eq:4}
	\end{split}
\end{equation}

The internal structure of the IFFE is shown in Fig. \ref{fig:3}(b). In \cite{38}, it was found that convolutional layers can serve as high-pass filters, reducing low-frequency information in features and increasing the variance and amplitude of high-frequency information. It has an amplifying effect on the high-frequency information. Therefore, in the IFFE, the input feature map \textbf{I} is passed through three convolutional layers sequentially to obtain outputs with different frequencies. Subsequently, the outputs of each convolutional layer are added together to obtain \textbf{I}\textsubscript{\textbf{1}} which contains all diversified frequency information obtained  \cite{26,27}. The number of convolutional layers is not fixed and it can be adjusted according to the actual situation. In this paper, three convolutional layers are selected by considering performance and the number of parameters, as shown in Eq. \ref{eq:5}.
\begin{equation}
	\begin{split}
		\label{eq:5}
		\mathrm{\textbf{I}}_{\textbf{1}} & =  {\mathrm{Conv}}_{{3} \times {3}}(\mathrm{\textbf{I}}) + {\mathrm{Conv}}_{{3} \times {3}}({\mathrm{Conv}}_{{3} \times {3}}(\mathrm{\textbf{I}}))  \\
		& + {\mathrm{Conv}}_{{3} \times {3}}({\mathrm{Conv}}_{{3} \times {3}}({\mathrm{Conv}}_{{3} \times {3}}(\mathrm{\textbf{I}})))
	\end{split}
\end{equation}

Then channel attention (CA) \cite{19} is utilized to adjust the weights of each channel and spatial attention (SA) \cite{44} is used in parallel to enhance the perception of critical information. Afterward, by concatenating their outputs, we can obtain beneficial frequency information \textbf{I}\textsubscript{\textbf{2}}, as shown in Eq. \ref{eq:6}. GAP is global average pooling, and GMP is global max pooling \cite{43}. Average pooling and max pooling can be used to aggregate the spatial information of feature maps \cite{48}.
\begin{equation}
	\label{eq:6}
 \mathrm{\textbf{I}}_{\textbf{2}}= \mathrm{Concat}[\mathrm{CA}(\mathrm{\textbf{I}}_{\textbf{1}}),\mathrm{SA}({\mathrm{\textbf{I}}_{\textbf{1}}})]
\end{equation}

Finally, a convolutional layer reduces the dimensionality of \textbf{I}\textsubscript{\textbf{2}} and then its output is added to M(\textbf{I}) to get the final output \textbf{O}, as in Eq. \ref{eq:7}
\begin{equation}
	\label{eq:7}
 \mathrm{\textbf{O}}= {\mathrm{Conv}}_{{1} \times {1}}(\mathrm{\textbf{I}}_{\textbf{2}})+\mathrm{M}(\mathrm{\textbf{I}}) 
\end{equation}

\subsection{Adaptive Learning Module}
\begin{figure}
	\includegraphics[height=3.2cm]{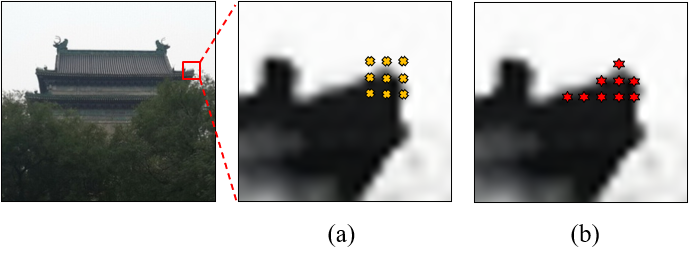}
	\caption{(a) Sampling points of regular convolution (b) Sampling points of deformable convolution}
	\label{fig:4}
\end{figure}

In most convolution operations, the shape of the sampling point is fixed, as shown in Fig \ref{fig:4}(a). This situation typically damages the texture details of objects, resulting in excessively smooth artifacts and making it challenging for networks to learn useful features in real-world scenes \cite{24}.

The introduction of deformable convolution \cite{31} can effectively address this issue. Compared to regular convolution, deformable convolution learns the offset of the convolution kernel, so that the sampling points of the original regular convolution kernel are shifted, as shown in Fig \ref{fig:4}(b). The offset can be dynamically adjusted during the training process to better fit the object itself. Therefore, we introduce the deformable convolution network (DCN) \cite{31} into the ALM to enhance the ability to restore texture details.
\begin{figure}
	\centering
	\includegraphics[height=6cm]{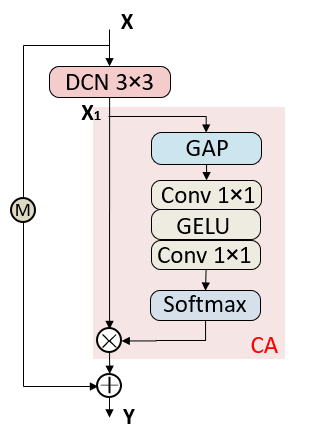}
	\caption{The internal structure of the ALM. DCN 3×3 is a deformable convolution with 9 sampling points}
	\label{fig:5}
\end{figure}

The structure of the ALM is shown in Fig. \ref{fig:5}. The input \textbf{X} undergoes DCN to obtain \textbf{X}\textsubscript{\textbf{1}}, and then a channel attention module is used to adjust the weight of each channel in \textbf{X}\textsubscript{\textbf{1}}. Finally, the output of CA is added to M(\textbf{X}) to obtain output \textbf{Y}, as in Eq. \ref{eq:8}.
\begin{equation}
	\label{eq:8}
	\mathrm{\textbf{Y}}=\mathrm{CA}(\mathrm{DCN}(\mathrm{\textbf{X}}))+\mathrm{M}(\mathrm{\textbf{X}})
\end{equation}
\begin{table*}[htbp]
	\centering
	\caption{Detailed architecture of the first-stage and the second-stage networks}
	\setlength{\tabcolsep}{1.7em}
	\begin{tabular}{p{8em}cccc}
		\toprule
		\multirow{2}{*}{Factor} & \multicolumn{2}{c}{TSNet-S} & \multicolumn{2}{c}{TSNet-L} \\
		\cmidrule(lr){2-3} \cmidrule(lr){4-5}
		& First stage & Second stage & First stage & Second stage \\
		\midrule
		Num of Blocks & [1, 1, 2, 1, 1] & [1, 2, 1] & [2, 2, 4, 2, 2] & [1, 2, 1] \\
		Embedding Dims & [24, 48, 96, 48, 24] & [24, 48, 24] & [24, 48, 96, 48, 24] & [24, 48, 24] \\
		\bottomrule
	\end{tabular}
	\label{tab:1} 
\end{table*}
\begin{table}[htbp]
	\centering
	\caption{The settings on different datasets}
	\begin{tabular}{l c c c}
		\toprule
		Datasets & Cropping size & Batch size & Epochs \\
		\midrule
		RESIDE-IN & 256×256 & 10    & 500 \\
		RESIDE-6K & 256×256 & 10    & 500 \\
		Haze-4K & 256×256 & 10    & 1000 \\
		Dense-Haze & 640×640 & 2     & 800 \\
		NH-Haze & 560×560 & 2     & 800 \\
		\bottomrule	
	\end{tabular}%
	\label{tab:2}
\end{table}%

The ALM are deployed before the first upsampling in both the first-stage and second-stage networks. This allows the ALM to understand the preceding features comprehensively while providing a well-prepared input for the subsequent network. In each stage, two ALM are utilized for recovering textures more efficiently while considering the number of parameters.

\subsection{Two-stage Design}
Unlike previous single-stage dehazing networks, we have added a network for optimizing dehazed images after the dehazing network, serving as the second-stage network of TSNet, as shown in Fig. \ref{fig:2}. We observed that after the hazy image passes through the first-stage network, the initial haze has been mostly removed. However, some areas still exhibit noticeable artifacts and color distortion. So we proposed to add the second-stage network to optimize the dehazed images obtained from the first-stage network.
\begin{table*}[htbp]
	\renewcommand{\arraystretch}{1.5}
	\centering
	\caption{Quantitative comparison of various methods on the RESIDE-IN, Haze-4K, and RESIDE-6K datasets}
	\begin{tabular}{p{15.5em} *{6}{c} *{2}{>{\centering\arraybackslash}m{4em}}}
		\toprule
		\multirow{2}[4]{*}{{Methods}} & \multicolumn{2}{c}{{RESIDE-IN}} & \multicolumn{2}{c}{{Haze-4K}} & \multicolumn{2}{c}{{RESIDE-6K}} &\multicolumn{2}{c}{Overhead} \\
		\cmidrule(lr){2-3} \cmidrule(lr){4-5} \cmidrule(lr){6-7} \cmidrule(lr){8-9}
		& {PSNR} & {SSIM} & {PSNR} & {SSIM} & {PSNR} & {SSIM} & \#Param & MACs \\
		\midrule
		DCP \cite{4}(TPAMI’10) & 16.62 & 0.818 & 14.01 & 0.76  & 17.88 & 0.816 & -     & - \\
		DehazeNet \cite{14}(TIP’16) & 19.82 & 0.821 & 19.12 & 0.84  & 21.02 & 0.87  & 0.009M  & 0.581G \\
		MSCNN \cite{12}(ECCV’16) & 19.84 & 0.833 & 14.01 & 0.51  & 20.31 & 0.863 & 0.008M  & 0.525G \\
		AOD-Net \cite{13}(WACV’17) & 20.51 & 0.816 & 17.15 & 0.83  & 20.27 & 0.855 & 0.002M  & 0.115G \\
		GridDehazeNet \cite{28}(ICCV’19) & 32.16 & 0.984 &   -    &   -    & 25.86 & 0.944 & 0.956M  & 21.49G \\
		MSBDN \cite{18}(CVPR’20) & 33.67 & 0.985 & 22.99 & 0.85  & 28.56 & 0.966 & 31.35M & 41.54G \\
		FFA-Net \cite{19}(AAAI’20) & 36.39 & 0.989 & 26.96 & 0.95  & 29.96 & 0.973 & 4.456M  & 287.8G \\
		AECR-Net \cite{24}(CVPR’21) & 37.17 & 0.99  & - & - & 28.52 & 0.964 & 2.611M  & 52.20G \\
		MAXIM \cite{20}(CVPR’22) & 38.11 & 0.991 &   -    &  -     & - & - & 14.1M & 216G \\
		PMNet \cite{23}(ECCV’22) & 38.41 & 0.99  & 33.49 & 0.98  &  -     &   -    & 18.9M & 81.13G \\
		DehazeFormer-M \cite{21}(TIP’23) & 38.46 & 0.994 & 31.82 & 0.985 & 30.89 & \textbf{0.977} & 4.634M  & 48.64G \\
		MixDehazeNet-B \cite{25}(CVPR’23) & 40.9  & \textbf{0.996} & -      &  -     & 30.69 & 0.974 & 6.25M & 43.61G \\
		\midrule
		TSNet-S & 41.07 & 0.995 & 33.74 & 0.989 & 30.77 & 0.974 & 2.486M & 30.06G \\
		TSNet-L & \textbf{42.6} & 0.995 & \textbf{34.95} & \textbf{0.991} & \textbf{31.31} & 0.975 & 4.366M & 47.01G \\
		\bottomrule	
	\end{tabular}%
	\label{tab:3}
\end{table*}

To keep TSNet as concise as possible, the second-stage network is simplified based on the first-stage network, and its structure can be considered a reduced version of the first-stage network. We introduced a loss function between the two stages. The hazy image undergoes processing by the first-stage network, resulting in a dehazed image. Then the dehazed image is utilized as the input for the second-stage network, which is responsible for optimizing the dehazed image. The two stages of TSNet are trained independently, and their weights are saved separately. 

\subsection{Learning Objective and Training Loss}
A ground truth image can be obtained by adding a hazy image to an opposite fog map, as shown by the dashed box in Fig. \ref{fig:1}. Previous networks directly learned ground truth images and output dehazed images. However, the differences between hazy and ground truth images are diverse, including saturation, sharpness, contrast, and so on. So it is challenging for networks to fully consider these differences during the learning process of ground truth images. 

Unlike previous methods, we change the learning objective of the first-stage network of TSNet from a ground truth image to an opposite fog image. TSNet outputs a learned opposite fog map b, which is then added to the hazy image to obtain the dehazed image c. This change allows the network to learn the differences between hazy images and ground truth images directly, which enhances its ability to learn the features of various haze types and improves the efficiency of network learning. We abbreviate the above method as CL. The second-stage network optimizes dehazed images by directly outputting the final dehazed images.

Assuming \textit{J} is the ground truth image in the dataset, and \textit{I} is the corresponding hazy image. \textit{F}\textsubscript{\textit{1}}(\textit{I}) is the output of the first-stage network in TSNet, and \textit{F}\textsubscript{\textit{2}}(\textit{I}) is the output of the second-stage network. Then the dehazed image of the first-stage network can be obtained by adding \textit{F}\textsubscript{\textit{1}}(\textit{I}) to \textit{I}.

The loss function $\mathcal{L}_{\text{1}}$ in the first stage consists of $\mathcal{L}_{\text{a}}$ and contrast loss $\mathcal{L}_{\text{b}}$:
\begin{equation}
\mathcal{L}_{\text {a}} = \mathrm{smooth\,L_1}(J,I + F_{1}(I))
\end{equation}
\begin{equation}
\label{eq:10}
\mathcal{L}_{\text {b}} = \beta \sum_{i=0}^n \omega_i \cdot \frac{\mathrm{L_1}\left(R_i(J), R_i(F_1(I)+I)\right)}{\mathrm{L_1}\left(R_i(I),  R_i(F_1(I)+I)\right)}
\end{equation} 
Where $\beta$ is a hyperparameter used to balance the contrastive learning loss and $\mathrm{smooth\,L_1}$ \cite{45}, $\mathrm{L_1}$(\textit{x}, \textit{y}) is the \textit{L}\textsubscript{{1}} loss, $\omega$\textsubscript{\textit{i}} is the weight coefficient, \textit{R}\textsubscript{\textit{i}} (\textit{i} = 1,2, ...,\textit{n}) represents extracting features from the fixed pre-trained ResNet-152 \cite{40} model at the \textit{i}-th layer. Then the loss function $\mathcal{L}_{\text{1}}$ can be expressed as:
\begin{equation}
\mathcal{L}_{\text {1}} = \mathcal{L}_{\text {a}}+\mathcal{L}_{\text {b}}
\end{equation}

The loss function $\mathcal{L}_{\text{2}}$ for the second-stage network can be expressed as:
\begin{equation}
\mathcal{L}_{\text {2}}=\mathrm{smooth\,L_1}(J, F_{2}(I))
\end{equation}

\section{Experiments}
\subsection{Datasets}
We evaluated our method on publicly available datasets RESIDE-IN, RESIDE-6K \cite{33}, and Haze-4K \cite{34}, as well as NH-Haze \cite{35} and Dense-Haze \cite{36}. RSIDE-IN contains 13390 pairs of synthetic indoor images for training and the test set is taken from 500 pairs of indoor images in SOTS. RESIDE-6K and Haze-4K are mixed datasets containing both indoor and outdoor synthetic images. Specifically, RESIDE-6K contains 3,000 pairs of indoor images and 3,000 pairs of outdoor images, with 5,000 pairs used for training and 1,000 pairs for testing. Haze-4k contains 4,000 pairs of synthetic images, with 3,000 pairs used for training and 1,000 pairs for testing. NH-Haze and Dense-Haze datasets consist of hazy images generated by a haze machine and their corresponding ground truth images. NH-Haze represents an uneven haze dataset, whereas Dense-Haze represents a dense haze dataset.

\subsection{Implementation Details}
We trained our network using one NVIDIA RTX-4090. The optimizer utilized AdamW \cite{41}, with exponential decay rates $\beta$\textsubscript{\textit{1}} and $\beta$\textsubscript{\textit{2}} set to 0.9 and 0.999 respectively. The learning rate used the cosine annealing strategy \cite{42} to reduce the initial learning rate from 4×10\textsuperscript{-4} to 4×10\textsuperscript{-2} during training. As the input and output dimensions of the multi-scale fusion module are the same, it can be used continuously at the same position in TSNet. Therefore, we provided two variants, TSNet-S and TSNet-L, with specific settings as shown in Table \ref{tab:1}. The weight $\beta$ of the contrastive loss was set to 0.1. We set $\mathrm{L_1}$(\textit{x}, \textit{y}) in Eq. \ref{eq:10} after the latent features of the 11th, 35th, and 143rd layers from the fixed pre-trained ResNet-152, with corresponding coefficients $\omega$\textsubscript{\textit{i}} (\textit{i} = 1,2,3) were set to 1, 1/4, and 1/8.

The size and number of images on different datasets are different, so the settings were different on each dataset \cite{39}, as shown in Table \ref{tab:2}.

\begin{figure*}
	\centering
	\includegraphics[width=1\textwidth]{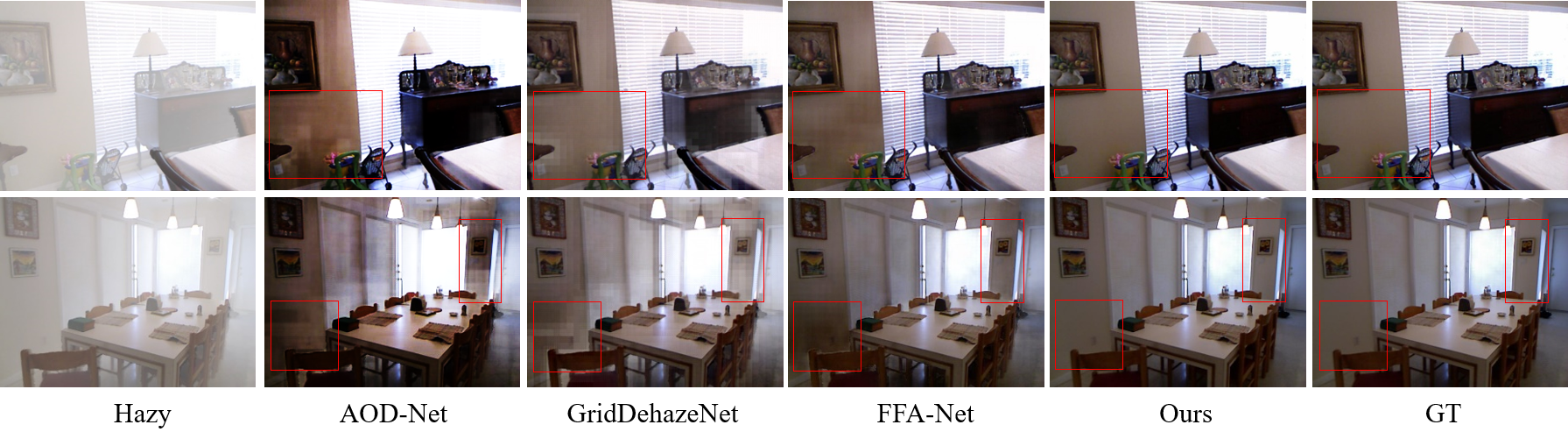}
	\caption{Qualitative comparison of various methods on the RESIDE-IN dataset. The first column is the hazy images and the last column is the ground truth images}
	\label{fig:6}
\end{figure*}
\begin{figure*}
	\centering
	\includegraphics[width=1.015\textwidth]{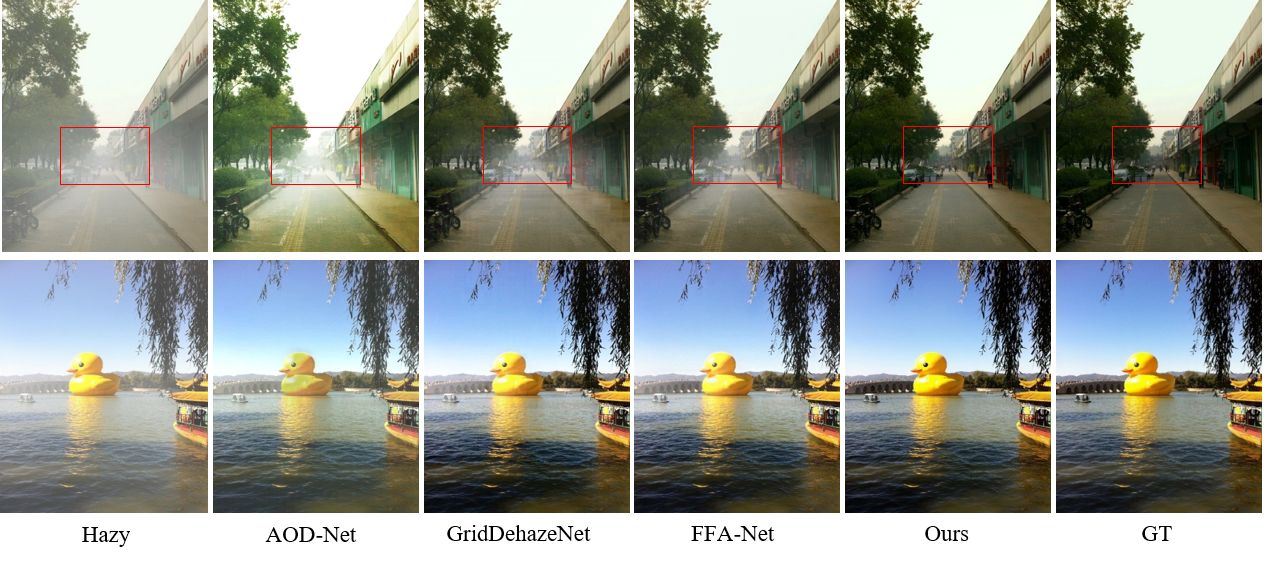}
	\caption{Qualitative comparison of various methods on the RESIDE-6K dataset}
	\label{fig:7}
\end{figure*}

\subsection{Comparisons with state-of-the-art methods}
We perform both quantitative and qualitative analyses on synthetic and real-world datasets, respectively.

\textbf{Synthetic datasets}: In Table \ref{tab:3}, we summarize the results on the three synthetic datasets. The data in bold represents the best performance on the corresponding metrics. The performance of TSNet improves as the number of the multi-scale fusion modules increases, as shown by the results of the two variants, TSNet-S and TSNet-L. Compared to most previous methods, TSNet-S has fewer parameters and computational costs. However, its performance is better. TSNet-L achieved the best performance on Haze-4K with 34.95dB PSNR and 0.991 SSIM compared to previous methods. Compared to the second-best-performing method, PMNet \cite{23}, TSNet-L showed a 4.5\% increase in PSNR while reducing the number of parameters by 76.9\%. Additionally, TSNet-L achieved the highest PSNR (42.60dB) on RESIDE-IN and the highest PSNR (31.31dB) on RESIDE-6K. Its SSIM (0.995) on RESIDE-IN is very close to the highest SSIM (0.996), and SSIM (0.975) on RESIDE-6K is very close to the highest SSIM (0.977). TSNet-S is suitable for tasks with higher demands for real-time performance, while TSNet-L can be applied to tasks with higher requirements for clarity.
\begin{table}[htbp]
	\centering
	\caption{Quantitative comparison of various methods on the NH-Haze and Dense-Haze datasets}
	\begin{tabular}{l cc cc}
		\toprule
		\multirow{2}[4]{*}{Methods} & \multicolumn{2}{c}{NH-Haze} & \multicolumn{2}{c}{Dense-Haze} \\
		\cmidrule(lr){2-3} \cmidrule(lr){4-5}
		& PSNR & SSIM & PSNR & SSIM \\
		\midrule
		DCP \cite{4} & 10.57 & 0.52  & 10.06 & 0.385 \\
		AOD-Net \cite{13} & 15.4  & 0.569 & 13.14 & 0.414 \\
		FFA-Net \cite{19} & 19.87 & 0.692 & 14.39 & 0.452 \\
		AECR-Net \cite{24} & 19.88 & 0.717 & 15.8  & 0.466 \\
		DehazeFormer-M \cite{21} & 18.59 & 0.754 & 16.13 & 0.549 \\
		MixDehazeNet-B \cite{25} & 17.17 & 0.718 & 14.89 & 0.567 \\
		\midrule
		TSNet-S & 18.5  & 0.756 & 16.74 & 0.613 \\
		TSNet-L & \textbf{19.94} & \textbf{0.799} & \textbf{17.48} & \textbf{0.639} \\
		\bottomrule
	\end{tabular}%
	\label{tab:4}%
\end{table}%
\begin{figure*}
	\centering
	\includegraphics[width=1.015\textwidth]{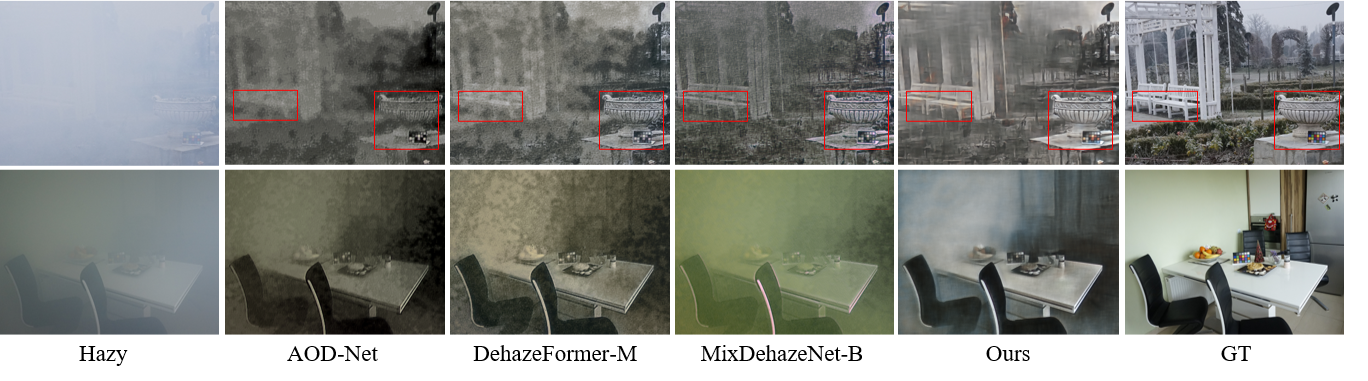}
	\caption{Qualitative comparison of various methods on the Dense-HAZE dataset. The first column is the hazy images and the last column is the ground truth images}
	\label{fig:8}
\end{figure*}
\begin{figure*}
	\centering
	\includegraphics[width=1\textwidth]{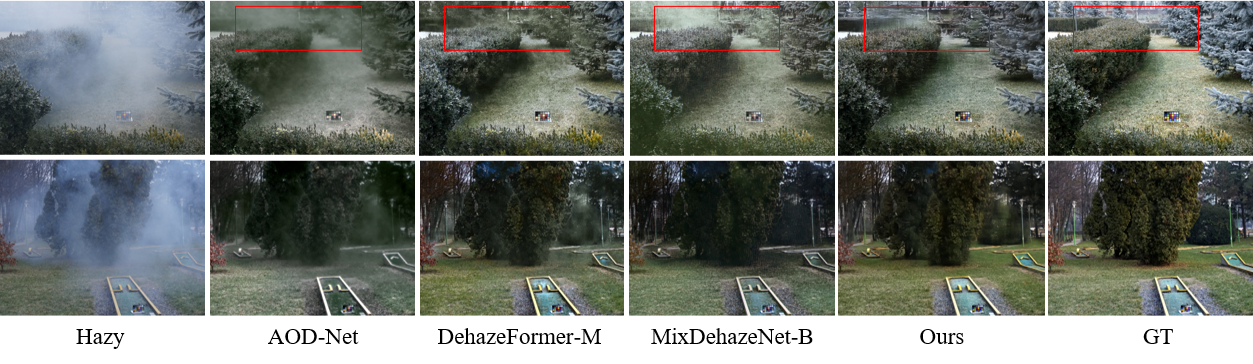}
	\caption{Qualitative comparison of various methods on the NH-HAZE dataset}
	\label{fig:9}
\end{figure*}

Fig. \ref{fig:6} illustrates the dehazed images of TSNet and other models on the RESIDE-IN dataset. Some regions show varying degrees of detail loss and severe blocky artifacts in the images restored by AOD-Net \cite{13}, GridDehazeNet \cite{28}, and FFA-Net \cite{19}. AOD-Net has severe colour distortion. Fig. \ref{fig:7} illustrates the outputs on the RESIDE-6K dataset. AOD-Net, GridDehazeNet, and FFA-Net fail to effectively remove distant haze and exhibit color discrepancies compared to ground truth images. In contrast, TSNet effectively removes haze with fewer artifacts on both RESIDE-IN and RESIDE-6K datasets. Additionally, its color is closer to ground truth images.

\textbf{Real-world datasets}: Many previous state-of-the-art methods performed well on synthetic datasets but exhibited poor performance on real-world datasets. However, TSNet achieved satisfactory performance on real-world datasets as well. From the results in Table \ref{tab:4}, TSNet-L achieves the best performance on both Dense-Haze and NH-Haze datasets compared to previous methods. On the Dense-Haze dataset, TSNet-L significantly outperforms the other methods, with 17.48dB PSNR and 0.639dB SSIM. The results demonstrate that TSNet can efficiently remove uneven haze and dense haze.

On the Dense-Haze dataset, the output images of AOD-Net \cite{13}, DehazeFormer-M \cite{21}, and MixDehazeNet-B \cite{25} all exhibit severe artifacts, while the outputs of TSNet appear smoother and better at recovering object details, as shown in Fig. \ref{fig:8}. Fig. \ref{fig:9} shows the output images on the NH-HAZE dataset. Many details are lost in the AOD-Net outputs, and residual haze remains in the DehazeFormer-M and MixDehazeNet-B outputs. In contrast, the outputs of TSNet effectively remove haze and restore details such as tree leaves and grass more comprehensively.
\begin{figure*}
	\centering
	\includegraphics[width=1\textwidth]{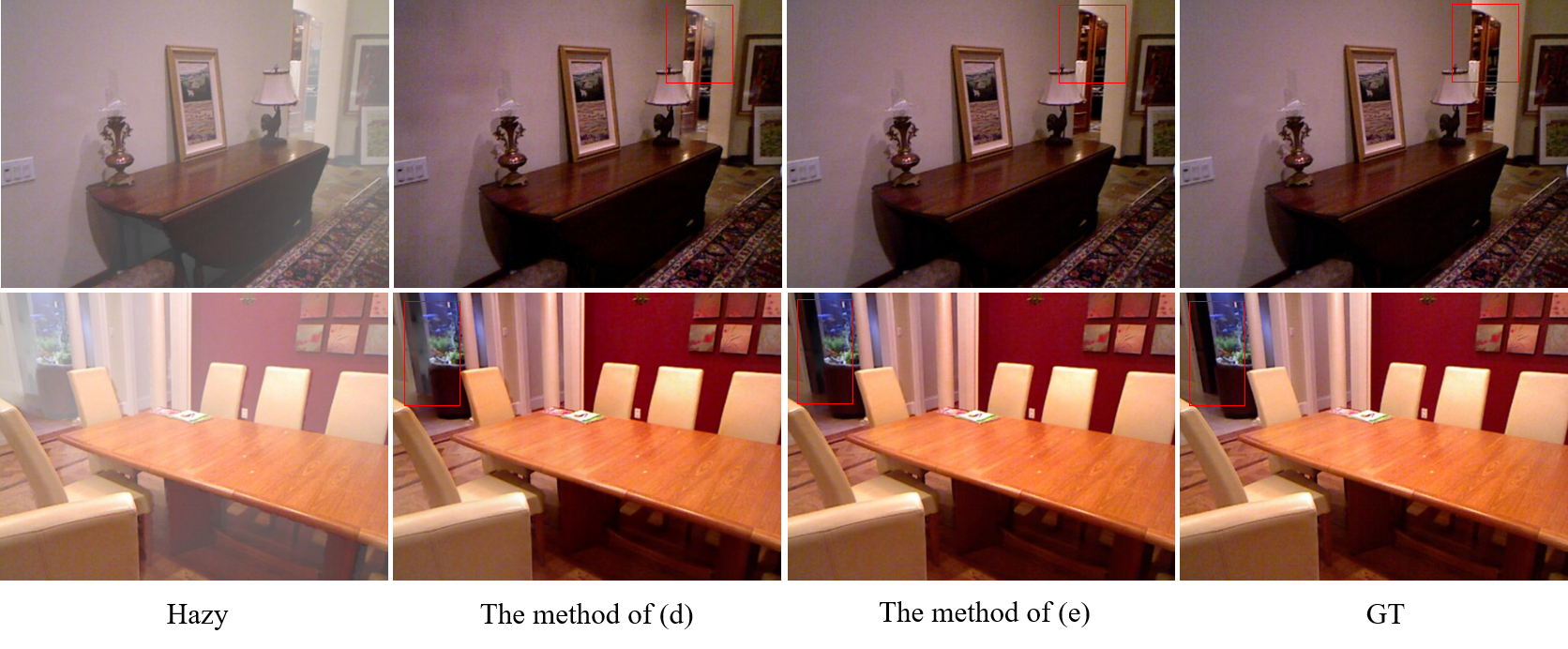}
	\caption{Qualitative comparison of the results obtained by methods (d) and (e)}
	\label{fig:10}
\end{figure*}
\begin{table}[htbp]
	\centering
	\caption{Ablation study on the RESIDE-IN and REISDE-6K datasets}
	\setlength{\tabcolsep}{4.2pt}
	\begin{tabular}{p{0.6em}p{9.8em} cc cc}
		\toprule
		\multicolumn{2}{c}{\multirow{2}[4]{*}{{Method}}} & \multicolumn{2}{c}{RESIDE-IN} & \multicolumn{2}{c}{RESIDE-6K} \\
		\cmidrule(lr){3-4} \cmidrule(lr){5-6}
		\multicolumn{2}{c}{} & \multicolumn{1}{c}{PSNR} & \multicolumn{1}{c}{SSIM} & \multicolumn{1}{c}{PSNR} & \multicolumn{1}{c}{SSIM} \\
		\midrule
		\multicolumn{1}{c}{} & Base  & 38.73 & 0.994 & 29.37 & 0.957 \\
		(a)   & Base+CL & 39.16 & 0.994 & 29.67 & 0.959 \\
		(b)   & Base+ALM & 39.2  & 0.993 & 29.75 & 0.966 \\
		(c)   & Base+ISSN & 39.6  & 0.993 & 30.2  & 0.972 \\
		(d)   & Base+CL+ALM & 39.89 & 0.994 & 29.96 & 0.966 \\
		(e)   & Base+CL+ALM+ISSN & 41.07 & 0.995 & 30.77 & 0.974 \\
		\bottomrule
	\end{tabular}%
	\label{tab:5}%
\end{table}%
\subsection{Ablation Study}
To demonstrate the effectiveness of the proposed methods, we conducted ablation experiments in this section. We performed the ablation experiments on the TSNet-S module with the following settings:

Base is the first-stage network of TSNet-S, but it does not include the ALM and the learning objective is ground truth images. (a) involves changing the learning objective from ground truth images to opposite fog maps in Base (CL). (b) adds the ALM to Base. (c) incorporates the second-stage network of TSNet-S into Base (ISSN), and the learning objective of the second-stage network is ground truth images. (d) represents changing the learning objective of the first-stage network (CL) while adding the ALM. (e) uses three methods simultaneously, which represents the complete TSNet-S. To ensure the rigor of the results, we conducted ablation experiments on two datasets, RESIDE-IN, and RESIDE-6K, with all networks trained in the same way.

In Table \ref{tab:5}, it can be concluded that the Base, which only adds the multi-scale fusion module, has already surpassed most of the previous state-of-the-art methods. Each of the method (a), (b), and (c) contributes to the improvement of the results, and the dehazing effect improves accordingly when using two or three methods simultaneously.

From Fig. \ref{fig:10}, it can be observed that the haze has been greatly eliminated in the outputs of the method (d), but there is residual haze and some color distortion in certain areas. However, the residual haze is further removed and the texture details are better restored in the outputs of the method (e). In addition, the outputs of the method (e) are closer in color to the ground truth images. (d) represents the first-stage network of TSNet-S, and (e) represents the complete TSNet-S. Therefore, the results demonstrate the second-stage network achieves optimization over the outputs of the first-stage network.

The method (c) is to add the second-stage network of TSNet-S to Base, so the original  network of Base is deepened. Typically, as the network becomes deeper, the amount of parameters also increase. Therefore, the dehazing effect may also be further improved.
\begin{table}[htbp]
	\centering
	\caption{Experimental investigation into the impact of two-stage design and changes in network depth on results}
	\setlength{\tabcolsep}{1em} 
	\begin{tabular}{p{9.5em} c c c}
		\toprule
		\multirow{2}[4]{*}{{Datasets}} & \multicolumn{1}{c}{Base*} & \multicolumn{1}{c}{TS-all} & \multicolumn{1}{c}{TSNet-S} \\
		
		\cmidrule(lr){2-2}\cmidrule(lr){3-3}\cmidrule(lr){4-4} 
		
		& \multicolumn{1}{c}{PSNR} & \multicolumn{1}{c}{PSNR} & \multicolumn{1}{c}{PSNR} \\
		\midrule
		Reside-in & 39.89 & 39.64 & 41.07 \\
		Reside-6k & 29.96 & 30.16 & 30.77 \\
		NH-HAZE & 17.5  & 17.83 & 18.5 \\
		Dense-HAZE & 16.41 & 16.44 & 16.74 \\
		\bottomrule
	\end{tabular}%
	\label{tab:6}%
\end{table}%
\begin{table}[htbp]
	\centering
	\caption{Comparison of results on the RESIDE-IN dataset after adding ALM to other methods}
	\setlength{\tabcolsep}{1.5em} 
	\begin{tabular}{p{12.5em} cc}
		\toprule
		\multirow{2}[4]{*}{{Methods}} & \multicolumn{2}{c}{RESIDE-IN} \\
		\cmidrule{2-3}    
		& {PSNR} & {SSIM} \\
		\midrule
		AECR-Net & 37.17 & 0.99 \\
		AECR-Net+ALM & 37.9  & 0.992 \\
		DehazeFormer-M & 38.46 & 0.994 \\
		DehazeFormer-M+ALM & 38.94 & 0.995 \\
		MixDehazeNet-B & 40.9  & 0.996 \\
		MixDehazeNet-B+ALM & 41.33 & 0.996 \\
		\bottomrule
	\end{tabular}%
	\label{tab:7}%
\end{table}%

We conducted additional experiments to demonstrate that the improvement in results from our two-stage network design was not solely due to the deeper network. Specifically, we connected the two stages of the network without setting the loss function in the middle, making it a whole network named TS-all. We trained it in the same way as before. Base* represents Base with the addition of methods (a) and (b).

Table \ref{tab:6} shows that the method of two-stage network design is more helpful for improving results on synthetic datasets or real-world datasets compared to TS-all. Therefore, this paper believes that when the network reaches a certain depth, adding loss functions at appropriate positions can regulate the learning process, enabling the network to have clear learning objectives at specific stages.

In addition, the ALM is easy to deploy because of its simple structure, so we conducted experiments to apply it to other networks. Table \ref{tab:7} shows that the ALM can be utilized as a universal module in other networks, enhancing the final dehazing performance of those networks.

\section{Conclusion}
In this paper, we propose a two-stage image dehazing network called TSNet, which mainly includes multi-scale fusion modules and adaptive learning modules. The multi-scale fusion module can obtain a large receptive field at multiple scales and it can integrate features at different frequencies to reduce the differences between inputs and learning objectives. The adaptive learning module can activate learning of regions of interest and exhibit strong capability in restoring the texture details. Therefore the generalization of TSNet can be improved by using the multi-scale fusion module and the adaptive learning module. Additionally, the two-stage network design can dehaze images and then optimize the dehazed images, reducing artifacts and color distortions in outputs. TSNet also changing the learning objective from clear images to the opposite fog maps to enhance the learning efficiency of the network. Experimental results demonstrated that TSNet surpasses previous state-of-the-art methods on both synthetic and real-world datasets, achieving excellent dehazing performance. In the future, we plan to work on a more lightweight network and implement real-time dehazing on edge devices.

\section*{Declarations}
\textbf{Conflict of interest} The authors declare that they have no known competing financial interests or personal relationships that could have appeared to influence the work reported in this paper.



\begin{thebibliography}{}

\bibitem{1}
Shen L, Zhao Y, Peng Q, et al. An iterative image dehazing method with polarization[J]. IEEE Transactions on Multimedia, 2018, 21(5): 1093-1107.
\bibitem{2}
Agrawal S C, Jalal A S. A comprehensive review on analysis and implementation of recent image dehazing methods[J]. Archives of Computational Methods in Engineering, 2022, 29(7): 4799-4850.
\bibitem{3}
McCartney E J. Optics of the atmosphere: scattering by molecules and particles[J]. New York, 1976.
\bibitem{4}
He K, Sun J, Tang X. Single image haze removal using dark channel prior[J]. IEEE transactions on pattern analysis and machine intelligence, 2010, 33(12): 2341-2353.
\bibitem{5}
He K, Sun J, Tang X. Guided image filtering[J]. IEEE transactions on pattern analysis and machine intelligence, 2012, 35(6): 1397-1409.
\bibitem{6}
Zhu Q, Mai J, Shao L. A fast single image haze removal algorithm using color attenuation prior[J]. IEEE transactions on image processing, 2015, 24(11): 3522-3533.
\bibitem{7}
Berman D, Avidan S. Non-local image dehazing[C]//Proceedings of the IEEE conference on computer vision and pattern recognition. 2016: 1674-1682. 
\bibitem{8}
Nazir A, Cheema M N, Sheng B, et al. OFF-eNET: An optimally fused fully end-to-end network for automatic dense volumetric 3D intracranial blood vessels segmentation[J]. IEEE Transactions on Image Processing, 2020, 29: 7192-7202. 
\bibitem{9}
Xie Z, Zhang W, Sheng B, et al. BaGFN: broad attentive graph fusion network for high-order feature interactions[J]. IEEE Transactions on Neural Networks and Learning Systems, 2021, 34(8): 4499-4513.
\bibitem{10}
Li J, Chen J, Sheng B, et al. Automatic detection and classification system of domestic waste via multimodel cascaded convolutional neural network[J]. IEEE transactions on industrial informatics, 2021, 18(1): 163-173.
\bibitem{11}
Sheng B, Li P, Ali R, et al. Improving video temporal consistency via broad learning system[J]. IEEE Transactions on Cybernetics, 2021, 52(7): 6662-6675.
\bibitem{12}
Nah S, Hyun Kim T, Mu Lee K. Deep multi-scale convolutional neural network for dynamic scene deblurring[C]//Proceedings of the IEEE conference on computer vision and pattern recognition. 2017: 3883-3891.
\bibitem{13}
Li B, Peng X, Wang Z, et al. An all-in-one network for dehazing and beyond[J]. arXiv preprint arXiv:1707.06543, 2017.
\bibitem{14}
Cai B, Xu X, Jia K, et al. Dehazenet: An end-to-end system for single image haze removal[J]. IEEE transactions on image processing, 2016, 25(11): 5187-5198.
\bibitem{15}
Chen Z, He Z, Lu Z M. DEA-Net: Single image dehazing based on detail-enhanced convolution and content-guided attention[J]. IEEE Transactions on Image Processing, 2024.
\bibitem{16}
Wu R Q, Duan Z P, Guo C L, et al. Ridcp: Revitalizing real image dehazing via high-quality codebook priors[C]//Proceedings of the IEEE/CVF Conference on Computer Vision and Pattern Recognition. 2023: 22282-22291.
\bibitem{17}
Wang Z, Jia J, Lyu P, et al. Efficient Dehazing with Recursive Gated Convolution in U-Net: A Novel Approach for Image Dehazing[J]. Journal of Imaging, 2023, 9(9): 183.
\bibitem{18}
Dong H, Pan J, Xiang L, et al. Multi-scale boosted dehazing network with dense feature fusion[C]//Proceedings of the IEEE/CVF conference on computer vision and pattern recognition. 2020: 2157-2167. 
\bibitem{19}
Qin X, Wang Z, Bai Y, et al. FFA-Net: Feature fusion attention network for single image dehazing[C]//Proceedings of the AAAI conference on artificial intelligence. 2020, 34(07): 11908-11915. 
\bibitem{20}
Tu Z, Talebi H, Zhang H, et al. Maxim: Multi-axis mlp for image processing[C]//Proceedings of the IEEE/CVF conference on computer vision and pattern recognition. 2022: 5769-5780. 
\bibitem{21}
Song Y, He Z, Qian H, et al. Vision transformers for single image dehazing[J]. IEEE Transactions on Image Processing, 2023, 32: 1927-1941.
\bibitem{22}
Guo Y, Gao Y, Liu W, et al. SCANet: Self-paced semi-curricular attention network for non-homogeneous image dehazing[C]//Proceedings of the IEEE/CVF Conference on Computer Vision and Pattern Recognition. 2023: 1884-1893.
\bibitem{23}
Ye T, Jiang M, Zhang Y, et al. Perceiving and modeling density is all you need for image dehazing[J]. arXiv preprint arXiv:2111.09733, 2021.
\bibitem{24}
Wu H, Qu Y, Lin S, et al. Contrastive learning for compact single image dehazing[C]//Proceedings of the IEEE/CVF conference on computer vision and pattern recognition. 2021: 10551-10560. 
\bibitem{25}
Lu L, Xiong Q, Chu D, et al. Mix Structure Block For Image Dehazing Network. arXiv 2023[J]. arXiv preprint arXiv:2305.17654.
\bibitem{26}
Jin Y, Sheng B, Li P, et al. Broad colorization[J]. IEEE transactions on neural networks and learning systems, 2020, 32(6): 2330-2343.
\bibitem{27}
Sheng B, Li P, Fang X, et al. Depth-aware motion deblurring using loopy belief propagation[J]. IEEE Transactions on Circuits and Systems for Video Technology, 2019, 30(4): 955-969.
\bibitem{28}
Liu X, Ma Y, Shi Z, et al. Griddehazenet: Attention-based multi-scale network for image dehazing[C]//Proceedings of the IEEE/CVF international conference on computer vision. 2019: 7314-7323.
\bibitem{29}
Liu X, Pedersen M, Wang R. Survey of natural image enhancement techniques: Classification, evaluation, challenges, and perspectives[J]. Digital Signal Processing, 2022, 127: 103547. 
\bibitem{30}
Liu Y, Yan Z, Chen S, et al. Nighthazeformer: Single nighttime haze removal using prior query transformer[C]//Proceedings of the 31st ACM International Conference on Multimedia. 2023: 4119-4128.
\bibitem{31}
Wang W, Dai J, Chen Z, et al. Internimage: Exploring large-scale vision foundation models with deformable convolutions[C]//Proceedings of the IEEE/CVF Conference on Computer Vision and Pattern Recognition. 2023: 14408-14419.
\bibitem{32}
Guo F, Yang J, Liu Z, et al. Haze removal for single image: A comprehensive review[J]. Neurocomputing, 2023.
\bibitem{33}
Li B, Ren W, Fu D, et al. Benchmarking single-image dehazing and beyond[J]. IEEE Transactions on Image Processing, 2018, 28(1): 492-505.
\bibitem{34}
Liu Y, Zhu L, Pei S, et al. From synthetic to real: Image dehazing collaborating with unlabeled real data[C]//Proceedings of the 29th ACM international conference on multimedia. 2021: 50-58.
\bibitem{35}
Ancuti C O, Ancuti C, Timofte R. NH-HAZE: An image dehazing benchmark with non-homogeneous hazy and haze-free images[C]//Proceedings of the IEEE/CVF conference on computer vision and pattern recognition workshops. 2020: 444-445.
\bibitem{36}
Ancuti C O, Ancuti C, Sbert M, et al. Dense-haze: A benchmark for image dehazing with dense-haze and haze-free images[C]//2019 IEEE international conference on image processing (ICIP). IEEE, 2019: 1014-1018.
\bibitem{37}
Huang L, Qin J, Zhou Y, et al. Normalization techniques in training dnns: Methodology, analysis and application[J]. IEEE transactions on pattern analysis and machine intelligence, 2023, 45(8): 10173-10196.
\bibitem{38}
Cui Y, Ren W, Cao X, et al. Image Restoration via Frequency Selection[J]. IEEE Transactions on Pattern Analysis and Machine Intelligence, 2023.
\bibitem{39}
Luo Z, Gustafsson F K, Zhao Z, et al. Refusion: Enabling large-size realistic image restoration with latent-space diffusion models[C]//Proceedings of the IEEE/CVF conference on computer vision and pattern recognition. 2023: 1680-1691.
\bibitem{40}
He K, Zhang X, Ren S, et al. Deep residual learning for image recognition[C]//Proceedings of the IEEE conference on computer vision and pattern recognition. 2016: 770-778.
\bibitem{41}
Loshchilov I, Hutter F. Decoupled weight decay regularization[J]. arXiv preprint arXiv:1711.05101, 2017.
\bibitem{42}
Loshchilov I, Hutter F. Sgdr: Stochastic gradient descent with warm restarts[J]. arXiv preprint arXiv:1608.03983, 2016.
\bibitem{43}
Lin M, Chen Q, Yan S. Network in network[J]. arXiv preprint arXiv:1312.4400, 2013.
\bibitem{44}
Chu X, Tian Z, Wang Y, et al. Twins: Revisiting the design of spatial attention in vision transformers[J]. Advances in neural information processing systems, 2021, 34: 9355-9366.
\bibitem{45}
Girshick R. Fast r-cnn[C]//Proceedings of the IEEE international conference on computer vision. 2015: 1440-1448.
\bibitem{46}
Ling P, Chen H, Tan X, et al. Single image dehazing using saturation line prior[J]. IEEE Transactions on Image Processing, 2023.
\bibitem{47}
Lin X, Sun S, Huang W, et al. EAPT: efficient attention pyramid transformer for image processing[J]. IEEE Transactions on Multimedia, 2021, 25: 50-61.
\bibitem{48}
Huang G, Wen Y, Qian B, et al. Attention-based multi-scale feature fusion network for myopia grading using optical coherence tomography images[J]. The Visual Computer, 2023: 1-12.






\end{thebibliography}


\end{document}